\documentclass[letterpaper]{article}
\usepackage{aaai25}  
\usepackage{aaai25}
\usepackage{times}
\usepackage{helvet}
\usepackage{courier}
\usepackage{graphicx} 
\usepackage{array}
\usepackage{amsmath}
\usepackage{graphicx}
\usepackage{multirow} 
\usepackage{float}
\usepackage{booktabs}
\usepackage{longtable}
\usepackage{booktabs}
\usepackage{multirow}
\usepackage{array}
\usepackage{caption}
\usepackage{color}
\usepackage{pifont}
\usepackage{xcolor} 
\usepackage{colortbl}
\usepackage{url}
\usepackage{aaai25}  
\usepackage{graphicx} 
\usepackage{natbib}  
\usepackage{caption} 
\usepackage{algorithm}
\usepackage{algorithmic}
\usepackage{newfloat}
\usepackage{listings}

\DeclareCaptionStyle{ruled}{labelfont=normalfont,labelsep=colon,strut=off} 
\floatstyle{ruled}
\newfloat{listing}{tb}{lst}{}
\floatname{listing}{Listing}
\begin{document}
%
\title{VProChart: Answering Chart Question Through Visual Perception Alignment Agent and Programmatic Solution Reasoning}
\author{
    Muye Huang\textsuperscript{\rm 1,2}, 
    Lingling Zhang\textsuperscript{\rm 1,2}\thanks{Corresponding author.}, 
    Lai Han\textsuperscript{\rm 1,2}, 
    Wenjun Wu\textsuperscript{\rm 1,3}, 
    Xinyu Zhang\textsuperscript{\rm 1,3}, 
    Jun Liu\textsuperscript{\rm 1,2}
}
\affiliations{
    \textsuperscript{\rm 1}School of Computer Science and Technology, Xi’an Jiaotong University\\
    \textsuperscript{\rm 2}MOE KLINNS Lab, Xi’an Jiaotong University\\
    \textsuperscript{\rm 3}Shaanxi Province Key Laboratory of Big Data Knowledge Engineering\\
    \{huangmuye, hanlai, nickjun98, zhang1393869716\}@stu.xjtu.edu.cn,\\
    \{zhanglling, liukeen\}@xjtu.edu.cn
}
\maketitle
\begin{abstract}
Charts are widely used for data visualization across various fields, including education, research, and business. Chart Question Answering (CQA) is an emerging task focused on the automatic interpretation and reasoning of data presented in charts. However, chart images are inherently difficult to interpret, and chart-related questions often involve complex logical and numerical reasoning, which hinders the performance of existing models. This paper introduces VProChart, a novel framework designed to address these challenges in CQA by integrating a lightweight Visual Perception Alignment Agent (VPAgent) and a Programmatic Solution Reasoning approach. VPAgent aligns and models chart elements based on principles of human visual perception, enhancing the understanding of chart context. The Programmatic Solution Reasoning approach leverages large language models (LLMs) to transform natural language reasoning questions into structured solution programs, facilitating precise numerical and logical reasoning. Extensive experiments on benchmark datasets such as ChartQA and PlotQA demonstrate that VProChart significantly outperforms existing methods, highlighting its ability to understand and reason with charts.
\end{abstract}
\begin{links}
    \link{Homepage}{https://github.com/MuyeHuang/VproChart}
\end{links}


\section{Introduction}
Chart is a widely used format for data visualization, prevalent in scientific papers and business reports. Chart Question Answering (CQA) aims to answer questions based on chart contexts, achieving the automatic analysis of data trends and the automatic generation of data reports. In recent years, the CQA task has garnered increasing attention in light of advancements in multi-modal understanding and reasoning techniques. \cite{DBLP:conf/cvpr/KaflePCK18,DBLP:conf/wacv/MethaniGKK20}

\begin{figure}
    \centering
    \includegraphics[width=1\linewidth]{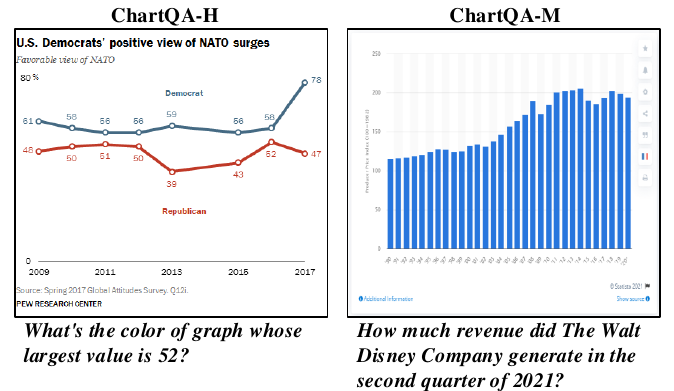}
    \caption{\small Comparison of the two subsets of ChartQA, ChartQA-H and ChartQA-M.}
    \label{DATASET}
\end{figure}

The CQA task naturally draws comparisons with Visual Question Answering (VQA), which involves answering questions based on images. Despite significant advances in VQA \cite{DBLP:journals/tip/LiHJSWC23,DBLP:journals/tip/WangWLZWW23}, CQA remains challenging due to the complex understanding of diagram contexts and complex reasoning requirements. Firstly, chart images are inherently difficult to interpret. For example, in the left chart of Figure \ref{DATASET}, the line chart does not have a strict legend, it only provides two labels corresponding to colors, which might potentially cause confusion. The elements and organizational methods in charts are not derived from nature but depend on the author's intentions. Different authors may use various elements or organizational methods for clarity or aesthetic appeal, complicating the understanding of chart images through common presentation rules. Secondly, questions in CQA often require complex numerical or logical reasoning. For example, in the left chart of Figure \ref{DATASET}, \textit{``Is the sum of the heights of the two highest red line greater than the value of the two highest blue line?"} necessitates numerical calculations and logical comparisons. These types of reasoning are particularly challenging for current research methods. However, existing approaches lack significant improvement in both of these aforementioned aspects.

On the one hand, the currently dominant strategy includes models like Pix2Struct \cite{DBLP:conf/icml/LeeJTH0EKSCT23} and UniChart \cite{DBLP:conf/emnlp/MasryKLHJ23}, which are end-to-end visual models, using extensive synthetic chart pretraining to enhance their chart understanding capabilities. This method has made progress in understanding synthetic charts but is still limited by the performance of external tools and synthetic data. On the other hand, MatCha \cite{DBLP:conf/acl/0001PKPLJACE23} and ChartInstruct \cite{DBLP:journals/corr/abs-2403-09028} have successfully enhanced the reasoning capabilities of models on CQA tasks through logical reasoning computation datasets or instruction fine-tuning. However, these reasoning approaches are still based on neural networks, making it difficult to achieve further progress in numerical reasoning. 
Overall, existing research lacks improvements for the two aforementioned challenges: in chart understanding, current methods rely on pretraining with large-scale chart-question pairs, while ignoring explicit modeling of relationships among chart elements; in numerical and logical reasoning, they rely on pretrained neural networks, which are not adept at handling complex logical and numerical reasoning.

To address the above two issues, we propose VProChart, an Agent-LLMs hybrid framework designed for the challenging CQA task. The VProChart method consists of two modules: the lightweight visual perception alignment agent and the programmatic solution reasoning approach. The Visual Perception Alignment agent (VPAgent) achieves chart understanding and basic chart question answering by principles of human visual perception. The programmatic solution reasoning approach converts questions into solution programs, which guide the reasoning process. During this process, the solution programs invokes the VPAgent to obtain intermediate variable values specified in the solution programs. Additionally, we perform a series of comparative experiments to demonstrate the superiority of VProChart.

Our main contributions can be summarized into four folds:
\begin{itemize}
    \item We propose the novel VProChart method to solve the challenges in understanding the context of the chart and the complex reasoning demands of the CQA task.
    \item We explicitly model and align the relationships between chart elements based on human visual perception principle, and infer answers within the modeled relationships using a question-driven reasoning module.
    \item We propose a programmatic solution reasoning framework, a flexible framework that can be applied to all types of questions, accurately inferring answers through flexible solution approaches and a precise Python-style reasoning engine.
    \item We conduct extensive experiments on different datasets such as ChartQA and PlotQA to validate the superiority of VProChart. The results show significant improvements compared to various popular competitors across different scales.
\end{itemize}

\section{Related Work}

\begin{figure*}[ht]
    \centering
    \includegraphics[width=1\linewidth]{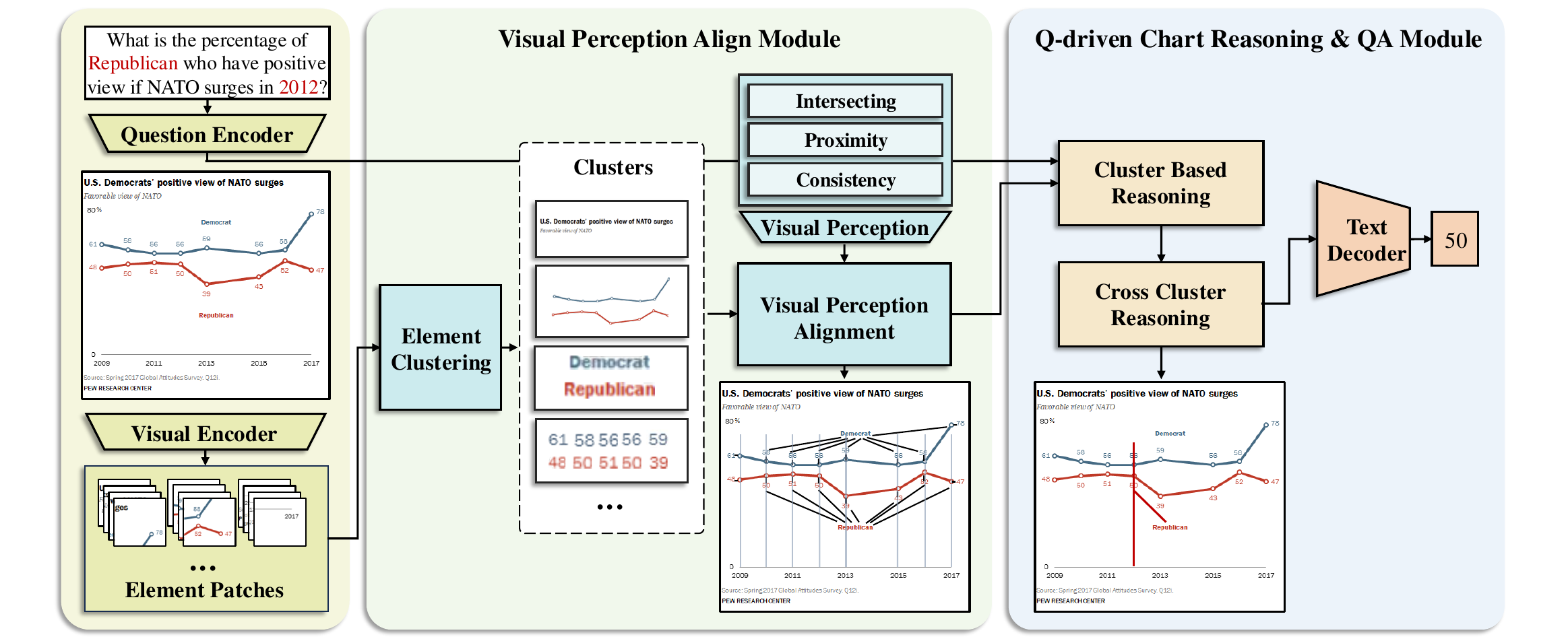}
    \caption{\small The overview of the proposed VPAgent. The proposed VPAgent comprises four modules: Chart \& Question Encoder Module (left), Visual Perception Align Module (center), Q-driven Chart Reasoning and Question Answering Module (right).}
    \label{overview vpa}
\end{figure*}

The chart question answering task answers textual questions by understanding visual context. Kahou et al. \cite{DBLP:conf/iclr/KahouMAKTB18} pioneered a synthetic Chart dataset FigureQA, which includes automatically generated chart images and a small number of template-generated true/false judgment questions. Kafle et al. \cite{DBLP:conf/cvpr/KaflePCK18} proposed another more comprehensive synthetic dataset, DVQA. They also introduced two models named MOM and SANDY to better handle CQA problems, and they subsequently introduced the PReFIL \cite{DBLP:conf/wacv/KafleSPCK20} model, achieving state-of-the-art results on FigureQA and DVQA. 

However, the aforementioned datasets lack the complex reasoning problems present in real-world CQA. Therefore, PlotQA \cite{DBLP:conf/wacv/MethaniGKK20} introduced a large-scale dataset, containing complex reasoning and numerical computation tasks. It comprises real-world data along with diverse synthetic charts. Furthermore, Masry et al. \cite{DBLP:conf/acl/MasryLTJH22} proposed the novel CQA dataset ChartQA, where the human subset consists of real-world images and manually curated questions, bringing the CQA task closer to real-world scenarios. CRCT \cite{DBLP:conf/eccv/LevyBL22} achieved satisfactory results on the PlotQA dataset by utilising a unified regression-classification head. Lee et al. \cite{DBLP:conf/icml/LeeJTH0EKSCT23} introduced a pure visual model Pix2Struct based on Vision Transformer \cite{DBLP:conf/iclr/DosovitskiyB0WZ21}, achieving state-of-the-art performance on various CQA datasets through large-scale pre-training on screenshots without the need for external OCR. Building upon this, MatCha \cite{DBLP:conf/acl/0001PKPLJACE23} using mathematical reasoning tasks during pre-training, resulting in significant improvements on ChartQA and PlotQA. Masry et al. \cite{DBLP:conf/emnlp/MasryKLHJ23} proposed UniChart, which achieved state-of-the-art performance on the ChartQA task through multi-task multi-stage pre-training. In light of the rapid advancement of LLMs, ChartLlama \cite{DBLP:journals/corr/abs-2311-16483} utilised Llama with an added visual branch, achieving good performance through large-scale pre-training and offering new directions for researchers. Moreover, ChartInstruct \cite{DBLP:journals/corr/abs-2403-09028}, ChartAst \cite{DBLP:journals/corr/abs-2401-02384} and TinyChart \cite{DBLP:journals/corr/abs-2404-16635}, an instructive approach to the CQA domain, which achieved new advancements in CQA tasks through the utilisation of LLMs.

The aforementioned works either rely on external detectors and data annotations or heavily on large-scale synthetic data driving, neglecting the exploration of structural relationships among chart elements and failing to achieve precise reasoning, limiting their understanding of charts and the reasoning capability for questions. In contrast, our work focuses on explicit modeling of chart structural relationships and precise question reasoning, achieving impressive results across multiple datasets.
\section{Method}
In this section, we introduce VProChart method, which comprises a lightweight visual perception alignment agent and programmatic solution reasoning approach. The programmatic solution reasoning approach transforms the reasoning problem into a programmatic solution and invokes the visual perception alignment agent to understand the chart context for reasoning execution.

\subsection{Visual Perception Alignment Agent}

The Visual Perception Alignment Agent (VPAgent) is designed for chart question answering. Figure\ref{overview vpa} illustrates the framework of the VPAgent, which includes four modules:
1) Chart \& question encoder module to extract sequences from charts and questions;
2) Visual perception align module to explicitly model the relationships between elements;
3) Q-driven chart reasoning module to infer and provide answers driven by the question;
4) Question Answering Module to produce the answer.
The details of these four modules are introduced as follows:

\subsubsection{1) Chart \& Question Encoder Module.}

The Chart \& Question Encoder Module converts the chart and question into sequential representations. This encoder is designed to recognize both textual and visual elements present $\mathbf{C_v}$ in the chart, and to represent the question in a sequential format that encapsulates the core essence $\mathbf{Q_\text{global}}$ of the question.

\paragraph{Chart Encoder Module.}
The encoding of chart images plays a critical role in the CQA process, as it requires the extraction of both visual and textual elements from the charts. To address the need for a lightweight and structurally complex encoder, we employ Swin-Encoder \cite{DBLP:conf/iccv/LiuL00W0LG21} as the chart encoder. Swin-Encoder is a lightweight encoder capable of achieving high-resolution encoding at a relatively low computational cost, enabling the simultaneous extraction of visual and textual information from images without the need for external OCR.
\begin{equation}
\mathbf{C_v = \{X_1, X_2, \ldots, X_n\}}
\end{equation}
where $\mathbf{C_v}$ is a sequence set containing chart information, and $\mathbf{X_k}$ represents the embedding of the $k$-th patch. 

\paragraph{Question Encoder Module.}
The Question Encoder Module encodes questions into sequences and extracts a global representation of each question. We employ RoBERTa \cite{DBLP:journals/corr/abs-1907-11692}, renowned for its state-of-the-art performance in natural language processing tasks, as the encoder. Given a question $\mathbf{Q}$, this procedure represents the question as a sequence:
\begin{equation}
\mathbf{Q_t}  =  \{\mathbf{H_0, H_1, H_2, \ldots, H_m}\}
\end{equation}
where $\mathbf{Q_t}$ is the sequential representation of question $\mathbf{Q}$, and $\mathbf{Q_\text{global}}$ is the global embedding of the question.

\subsubsection{2) Visual Perception Align Module.}

In the image sequence set $\mathbf{C_v}$, each patch contains elements of the chart. This module clusters elements of the same type within the image sequence. It then models the structural relationships among these elements by aligning elements from different clusters based on human visual perception principles. Finally, this module represents the visual perception alignment information as a matrix $\mathbf{V}$.

\paragraph{Element Clustering.}
Chart images come in various forms (\emph{e.g.}, pie charts, line charts, bar charts), each with distinct element names and features. To address this diversity of elements, we employ agglomerative hierarchical clustering, a flexible and noise-robust approach. This method operates at the patch level within chart images, organizing the differentiated elements into clusters:
\begin{equation}
\mathbf{G} = \left\{ \mathbf{G}_0, \mathbf{G}_1, \ldots, \mathbf{G}_{k-1} \right\}, \quad \left| \mathbf{G} \right| = k
\end{equation}
\begin{equation}
\mathbf{C}_v = \bigcup_{i=0}^{k-1} \mathbf{G}_i, \quad \mathbf{G}_i \cap \mathbf{G}_j = \emptyset \; \forall i \neq j
\end{equation}
where $k$ denotes the number of clusters, $\mathbf{G}$ is the set of all clusters, and $\mathbf{C_v}$ is the set of image embedding sequences.

\paragraph{Visual Perception Alignment.}
In this module, we align elements from different clusters based on visual perception principles:
 \textbf{a) Crosshair Intersecting Principle}, aligning elements based on their horizontal and vertical relationships on the original chart image, e.g., aligning the axes with the lines or bars. This is because charts always adhere to the Crosshair Intersecting alignment relationship for the x and y axes.
 \textbf{b) Spatial Proximity Principle}, alignment is determined by the spatial proximity of elements on the original chart, e.g., aligning bars with their corresponding numerical labels. This is because elements that are spatially proximate within the chart are more likely to be related. 
 \textbf{c) Color \& Semantic Consistency Principle}, alignment is based on the color and semantic similarity of elements, e.g., aligning green legend labels with green bars. This is because elements of the same color in the chart commonly represent the same attribute.

These three principles of alignment relationships can be respectively represented as $\mathbf{I}$, $\mathbf{P}$, and $\mathbf{S}$. 
For $\mathbf{I}$, we establish relationships with patches that are in the same row or column as patch $X_i$, and store the collection of these relationships within it.
For $\mathbf{P}$, we establish relationships with patches that are within a straight-line distance of 3 from patch $X_i$, and store the collection of these relationships within it. 
For $\mathbf{S}$, we calculate the normalized semantic similarity with patch $X_i$ and store these relationships within it.

The primary alignment method varies among different types of charts. For instance, pie charts predominantly employ spatial proximity, while line graphs primarily utilize crosshair intersecting. We utilize the global pooling representation $\mathbf{X_p}$ of the chart images to obtain the weights $\mathbf{W_c}$ of the three alignment through a MLP:
\begin{equation}
\mathbf{W}_c = \text{\textit{softmax}} \left( \mathbf{W}_2 \cdot \sigma \left( \mathbf{W}_1 \cdot \mathbf{X}_p + \mathbf{b}_1 \right) + \mathbf{b}_2 \right)
\end{equation}
However, not every pair of clusters in the chart needs to be aligned. The necessity of alignment between clusters is determined by computing the interactions among the global representations of each cluster. This ensures that the alignment process is effective, focusing only on the most relevant cluster interactions. Let $\mathbf{G}$ be the matrix of global representations of clusters:

\begin{equation}
\mathbf{G} = \left[ \mathbf{g}_0, \mathbf{g}_1, \mathbf{g}_2, \ldots, \mathbf{g}_{k-1} \right]
\end{equation}

The interaction matrix $\mathbf{W}$, which denotes the necessity scores of alignment between clusters, is computed as:
\begin{equation}
\mathbf{W} = \text{\textit{softmax}}(\mathbf{G} \cdot \mathbf{G}^T)
\end{equation}
By combining the weights $\mathbf{W}_c$ and $\mathbf{W}$, we integrate the alignment relationships between clusters into a visual criterion alignment matrix $\mathbf{V}$:
\begin{equation}
\mathbf{V} = \mathbf{W}_c \odot \left( \mathbf{W} \odot (\mathbf{I} + \mathbf{P} + \mathbf{S}) \right)
\end{equation}
Here, $\odot$ denotes the element-wise multiplication. Overall, the visual perception alignment module explicitly provides the alignment information $\mathbf{V}$ for chart image elements by integrating visual perception principles and multi-weight alignment information.

\subsubsection{3) Q-driven Chart Reasoning QA Module.}

Given a chart $\mathbf{C_v}$, this module mines implicit information within the graph and marks answers under the guidance of the problem-global representation $\mathbf{H_0}$ and the chart alignment information $\mathbf{V}$. To achieve this, we designed the intra-cluster reasoning module and the cross-cluster reasoning module to handle intra-cluster and inter-cluster reasoning tasks respectively. 

\paragraph{Intra-Cluster Reasoning.}
In charts, certain information such as omitted interval labels is implicitly provided. This module deeply mines the implicit information within clusters by using a restricted-attention mechanism. The mechanism confines the attention scope to the range of element clusters $\mathbf{G_i}$. It is followed by feed-forward layers, skip connections, and normalization. We stack this attention module for $\mathbf{N}$ layers to enhance information extraction.

\begin{equation}
\mathbf{C_r} = \sum_{i=0}^{k} \left[ \text{\textit{Softmax}}\left(\frac{\mathbf{W_Q C_v} [\mathbf{W_K G_i}]^T}{\sqrt{d_k}}\right) \mathbf{W_V G_i} \right]^N
\end{equation}

Here, $\mathbf{C_r}$ represents the output image sequence set of this module.

\paragraph{Cross-Cluster Reasoning.}
This module utilizes the alignment matrix \(\mathbf{V}\) and the global embedding of the question \(\mathbf{Q_\text{global}}\) to annotate elements in the alignment information \(\mathbf{V}\) that are strongly associated with the answer. Specifically, it computes the similarity scores between image elements in \(\mathbf{C_r}\) and the global representation \(\mathbf{Q_\text{global}}\) of the question. The top-k elements most relevant to the query are then selected and annotated in the alignment matrix \(\mathbf{V}\):
\begin{equation}
\mathbf{V_r} = \text{Annotate}\left(\text{\textit{softmax}}\left(\mathbf{C_r} \mathbf{Q_{\text{global}}^T}\right), \mathbf{V}\right)
\end{equation}

Here, \(\mathbf{V_r}\) represents the reasoned alignment matrix with the most relevant elements annotated. As the alignment matrix \(\mathbf{V}\) only contains alignment relationships between elements from different clusters, marking one element results in automatically marking elements from other clusters that are aligned with it. Consequently, the reasoned chart \(\mathbf{C_r}\) obtained from the previous steps is combined with the reasoned alignment information \(\mathbf{V_r}\), yielding the output \(\mathbf{O_c}\). This combination is achieved through a normalization process, which ensures the stability and consistency of the output:

\begin{equation}
\mathbf{O_c} = \frac{\mathbf{C_r} + \mathbf{V_r} - \mu}{\sqrt{\sigma^2 + \epsilon}} \gamma + \beta
\end{equation}

Here, \(\mu\) and \(\sigma^2\) represent the mean and variance of \(\mathbf{C_r} + \mathbf{V_r}\), respectively, while \(\epsilon\) is a small constant for numerical stability. The parameters \(\gamma\) and \(\beta\) are learnable affine transformation parameters.

\subsubsection{4) Question Answering Module.}
For a given question $\mathbf{Q}$, the answer can be either a floating-point number or textual content. Previous works, such as T5 \cite{DBLP:journals/jmlr/RaffelSRLNMZLL20}, have demonstrated that generative decoders can effectively handle both types of responses simultaneously. Consequently, we employ the mBART \cite{DBLP:journals/tacl/LiuGGLEGLZ20} decoder as the text decoder. The text prompt is provided as input to the decoder, which generates the answer based on the prompt context. 
The loss function for this process is defined as follows:
\begin{equation}
\mathcal{L} = -\sum_{t=1}^{T} \log p(y_t \mid y_{<t}, \mathbf{O_c}, \mathbf{Q})
\end{equation}
where \(y_t\) represents the target token at time step \(t\), \(y_{<t}\) denotes the sequence of previous target tokens, and \(p(y_t \mid y_{<t}, \mathbf{O_c}, \mathbf{Q})\) is the predicted probability of \(y_t\) given \(y_{<t}\), \(\mathbf{O_c}\), and \(\mathbf{Q}\). This loss function minimizes the negative log-likelihood of the predicted tokens.

\begin{figure}

    \centering
    \includegraphics[width=\linewidth]{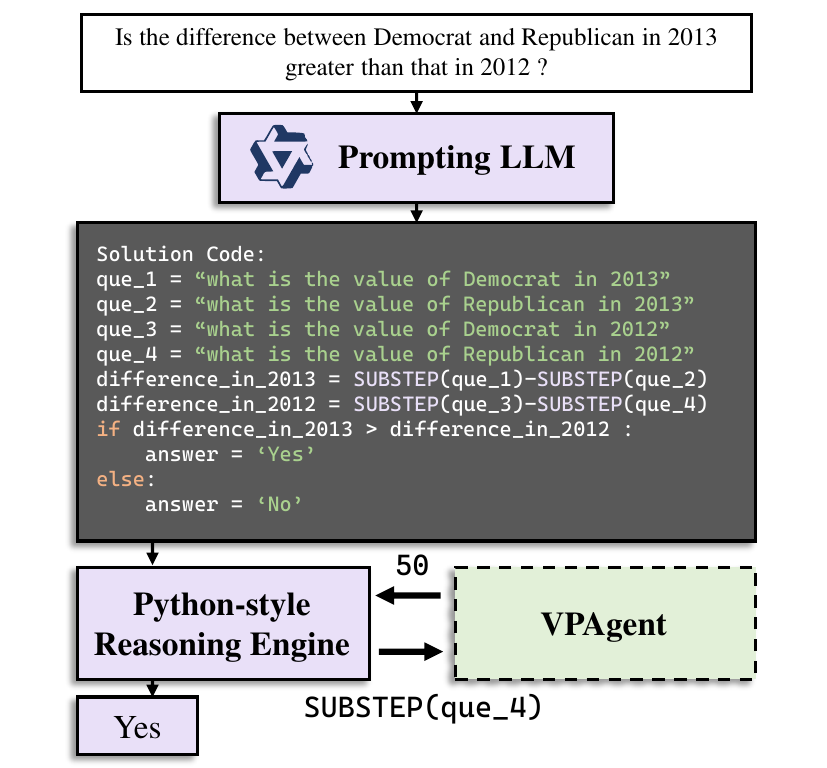}
    \caption{\small An example of Programmatic Solution Reasoning. }
    \label{programmatic}
\end{figure}

\subsection{Programmatic Solution Reasoning}
CQA necessitates advanced reasoning capabilities. To address this challenge, we propose the Programmatic Solution Reasoning approach, which leverages LLMs to integrate the precise reasoning of programs with the language understanding of LLMs. As depicted in Figure \ref{overview vpa}, the framework consists of two distinct modules:
1) Prompting LLMs to parse natural language questions into structured pseudo-code solution in Python style.
2) Python-style Reasoning Engine to interpret the solution generated by the LLMs and output answers through programming.

\noindent\textbf{1) Prompting LLMs.}

Prompting LLMs leverage the complex question understanding and code generation capabilities of LLMs to transform intricate textual questions into solution program \cite{xu2023symbol}. LLMs do not need to compute or infer answers, instead, they express the process of solving the problem in the form of solution program \cite{DBLP:journals/tmlr/ChenM0C23}. In particular, we prompt the LLMs to act as programming experts, solving problems programmatically by invoking two kinds of operators:
1) $ASK(\cdot)$. This operator takes the complete question as input and returns the answer directly, used for questions that do not require reasoning or computation. For instance, the question ``\textit{Which color indicates 65+ years?}" requires no reasoning.
2) $SUBSTEP(\cdot)$. This operator takes in sub-questions and returns answers to these sub-questions, used for complex questions requiring decomposition. 

\noindent\textbf{2) Python-style Reasoning Engine.}

Given any question \(\mathbf{Q}\) and its corresponding solution \(\mathbf{P}\), this module parses the solution program \(\mathbf{P}\) and employs the VPAgent to obtain values for critical variables within the program. The flexibility of the programming design endows the reasoning engine with broad applicability to various types of questions, ensuring robust performance across diverse query scenarios.
For example, as shown in Figure \ref{programmatic}, the key variable \textit{difference\_in\_2012} in the solution program requires calling the VPAgent twice to obtain the answer. The Reasoning Engine passes the variable ``what is the value of Republican in 2012" from the SUBSTEP to the VPAgent, and returns the output of the Agent, ``50." Finally, the Reasoning Engine uses a judgment statement to obtain the answer output of the solution program. This process ensures the accuracy of the reasoning.

\section{Experiments}

\begin{table*}[ht]
%
\centering
\begin{tabular}{>{\centering\arraybackslash}p{2.2cm} >{\centering\arraybackslash}p{1.1cm} *{3}{>{\centering\arraybackslash}p{1.1cm}} *{2}{>{\centering\arraybackslash}p{1.8cm}} *{3}{>{\centering\arraybackslash}p{1.1cm}}}
\toprule
\multirow{2}{*}{Model} & \multirow{2}{*}{Size} & \multicolumn{3}{c}{ChartQA} & \multicolumn{2}{c}{PlotQA} & \multicolumn{3}{c}{DVQA} \\ 
\cmidrule(lr){3-5} \cmidrule(lr){6-7} \cmidrule(lr){8-10}
                       &                       & Human   & Augment   & Avg   & \multicolumn{1}{>{\centering\arraybackslash}p{1.8cm}}{PlotQA-D1}    & \multicolumn{1}{>{\centering\arraybackslash}p{1.8cm}}{PlotQA-D2}   & Familiar & Novel & Avg   \\ 
\midrule
T5                     & 223M                  & 25.1    & 41.0      & 33.1  & 72.6        & 56.2       & 89.0    & 76.9 & 83.0 \\
PReFIL                 & -                     & -       & -         & -     & 57.9        & 10.4       & 80.9    & 80.0 & 80.5 \\
VisionTapas            & 223M                  & 29.6    & 61.4      & 45.5  & 65.3        & 42.5       & 94.4    & 95.5 & 94.9 \\
CRCT                   & -                     & -       & -         & -     & 76.9         & 34.4       & -        & -     & 82.1 \\
Pix2Struct             & 300M                  & 30.5    & 81.6      & 56.0  & 73.2         & 71.9        & -        & -     & - \\
ChartReader            & -                     & -       & -         & 52.6  & 78.1         & 59.3        &  \underline{95.4}     &  \underline{96.5}  &  \underline{96.0} \\
MatCha                 & 300M                  & 38.2    & \textbf{90.2}     & 64.2         & 92.3        & \textbf{90.7}     & -     & -     & - \\
UniChart               & 260M                  & 43.9    & 87.8      & 65.8  & 28.6         & 21.7        & -        & -     & - \\
\rowcolor{gray!20}VPAgent(Ours)          & 311M                  & \underline{48.1}& 87.4 &  \underline{67.8}&  \underline{92.6}  & 78.1  & \textbf{98.5} & \textbf{97.1}   & \textbf{97.8} \\
\rowcolor{gray!20}VProChart(Ours)         & 7B+0.3B               & \textbf{57.4}    &  \underline{88.0}      & \textbf{72.7}  & \textbf{94.7}            &  \underline{83.5}           & -        & -     & - \\
\bottomrule
\end{tabular}
\caption{Comparison results on ChartQA, PlotQA, and DVQA Datasets}
\label{mainresult}
\end{table*}

\subsection{Experimental Setting}
\noindent\textbf{Datasets.} 
We conducted extensive experiments on three prominent datasets: PlotQA \cite{DBLP:conf/wacv/MethaniGKK20}, DVQA \cite{DBLP:conf/cvpr/KaflePCK18}, and ChartQA \cite{DBLP:conf/acl/MasryLTJH22}. DVQA is a fully synthetic dataset focused on chart-specific questions. PlotQA, a large synthetic dataset, comprises two subsets, D1 and D2, incorporating both real-world data and automatically generated charts, with numerous numerical and logical reasoning questions. ChartQA includes two benchmarks for challenging CQA tasks: ChartQA-H and ChartQA-M. ChartQA-H features manually curated chart images and question-answer pairs, while ChartQA-M consists of machine-generated images and questions. As illustrated in Figure \ref{DATASET}, ChartQA-H poses a more formidable challenge due to its extensive collection of manually curated charts and complex multi-step reasoning questions.

\noindent\textbf{Competitors.} We evaluate our model against both fully-supervised models and large-language-model-based baselines. \textit{Fully-supervised models} include T5 \cite{DBLP:journals/jmlr/RaffelSRLNMZLL20}, PReFIL \cite{DBLP:conf/wacv/KafleSPCK20}, VisionTapas \cite{DBLP:conf/acl/MasryLTJH22}, Pix2Struct \cite{DBLP:conf/icml/LeeJTH0EKSCT23}, ChartReader \cite{DBLP:conf/iccv/ChengDH23}, CRCT \cite{DBLP:conf/eccv/LevyBL22}, MatCha \cite{DBLP:conf/acl/0001PKPLJACE23}, and UniChart \cite{DBLP:conf/emnlp/MasryKLHJ23}, all of which are trained on CQA benchmarks. \textit{Large-language-model-based models} such as PaLI-17B (res.588) \cite{DBLP:conf/iclr/Chen0CPPSGGMB0P23}, Qwen-VL \cite{Qwen-VL}, ChartLLama \cite{DBLP:journals/corr/abs-2311-16483}, OneChart \cite{DBLP:journals/corr/abs-2404-09987}, ChartGemma \cite{masry2024chartgemmavisualinstructiontuningchart} and ChartInstruct \cite{DBLP:journals/corr/abs-2403-09028} utilize end-to-end LLMs integrated with a visual component.

\noindent\textbf{Settings.} In our proposed VPAgent, we use the Donut weights \cite{DBLP:conf/eccv/KimHYNPYHYHP22} for initialization. The training process consists of two stages: firstly, pre-training 200k steps on the UniChart pretraining corpus \cite{DBLP:conf/emnlp/MasryKLHJ23}, with an initial learning rate of 1e-4; secondly, fine-tuning 5 epoch on the downstream dataset, with an initial learning rate of 5e-5. Notably, due to the substantial computational cost, we trained the model only on PlotQA-D1 and tested it on PlotQA-D2. In our programmatic solution reasoning framework, we employ \textit{Qwen2-7B-Instruct} \cite{qwen2} as the base LLM. All experiments were conducted using two Tesla A100 GPUs.

\subsection{Performance Comparison}
In Table \ref{mainresult} and \ref{chartqaresult}, we present a comprehensive summary of VProChart's performance on three popular datasets. From these results, we make the following three observations.

\begin{table}[ht]
\centering
\small
\begin{tabular}{p{2.4cm} >{\centering\arraybackslash}p{1.5cm} >{\centering\arraybackslash}p{0.8cm} >{\centering\arraybackslash}p{0.8cm} >{\centering\arraybackslash}p{0.8cm}}
\toprule
\multirow{2}{*}{Model} & \multirow{2}{*}{Size} & \multicolumn{3}{c}{ChartQA} \\ 
\cmidrule(lr){3-5}
                       &                       & Human   & Aug   & Avg   \\ 
\midrule
PaLI(res.588)          & 17B                   & 30.4    & 64.9      & 47.6  \\
Qwen-VL                & 9.6B                  & 44.3    & 78.9      & 61.6  \\
ChartLlama             & 13B                   & 48.9    &  \underline{90.4}      & 69.7  \\
ChartInstruct-E        & 7B                    & 45.5    & 87.8      & 66.7  \\
ChartInstruct-P        & 3B+0.3B                    &  \underline{50.1}    & \textbf{93.8}      &  \underline{71.9}  \\
OneChart               & 13B+0.2B                    & 49.1    & 85.3      & 67.2  \\
\rowcolor{gray!20}VProChart(Ours)         & 7B+0.3B               & \textbf{57.4}    & 88.0      & \textbf{72.7}  \\
\bottomrule
\end{tabular}
\caption{\small Comparison VLM results on ChartQA Dataset}
\label{chartqaresult}
\end{table}

1) Our VPAgent and VProChart method consistently outperform other competitors across three datasets in terms of overall performance. When comparing our VProChart method with similarly sized large models, the performance is nearly 5.0\% higher compared to the best competitors on ChartQA. Additionally, when the small VPAgent is used for basic chart question-answering tasks and compared with similarly sized fully supervised models, its performance is higher than the best competitor UniChart by 2\%. This indicates that our proposed VProChart method is advantageous for CQA tasks.

2) Compared to the best competitor ChartInstruct-P, our VProChart method achieves a performance improvement of over 7\% on ``human" set. Even without utilizing additional training data, the VPAgent surpasses the best competitor, UniChart, by more than 4\% on the human subset. It's worth noting that the gap between the small VPAgent and ChartLlama on the human subset is less than 1\%. This illustrates that the proposed VProChart approach exhibits strong generalization and is suitable for real-world CQA tasks.

3) With all methods, the performance scores worsen when the benchmark changes from augment to human, or from PlotQA-D1 to D2. This is reasonable because the charts in the human subset are manually curated, and the questions entail multi-step complex reasoning. Although PlotQA-D2 lacks manually curated charts, most questions still require multi-step reasoning.

\begin{figure*}[ht]
    \centering
    \includegraphics[width=1\linewidth]{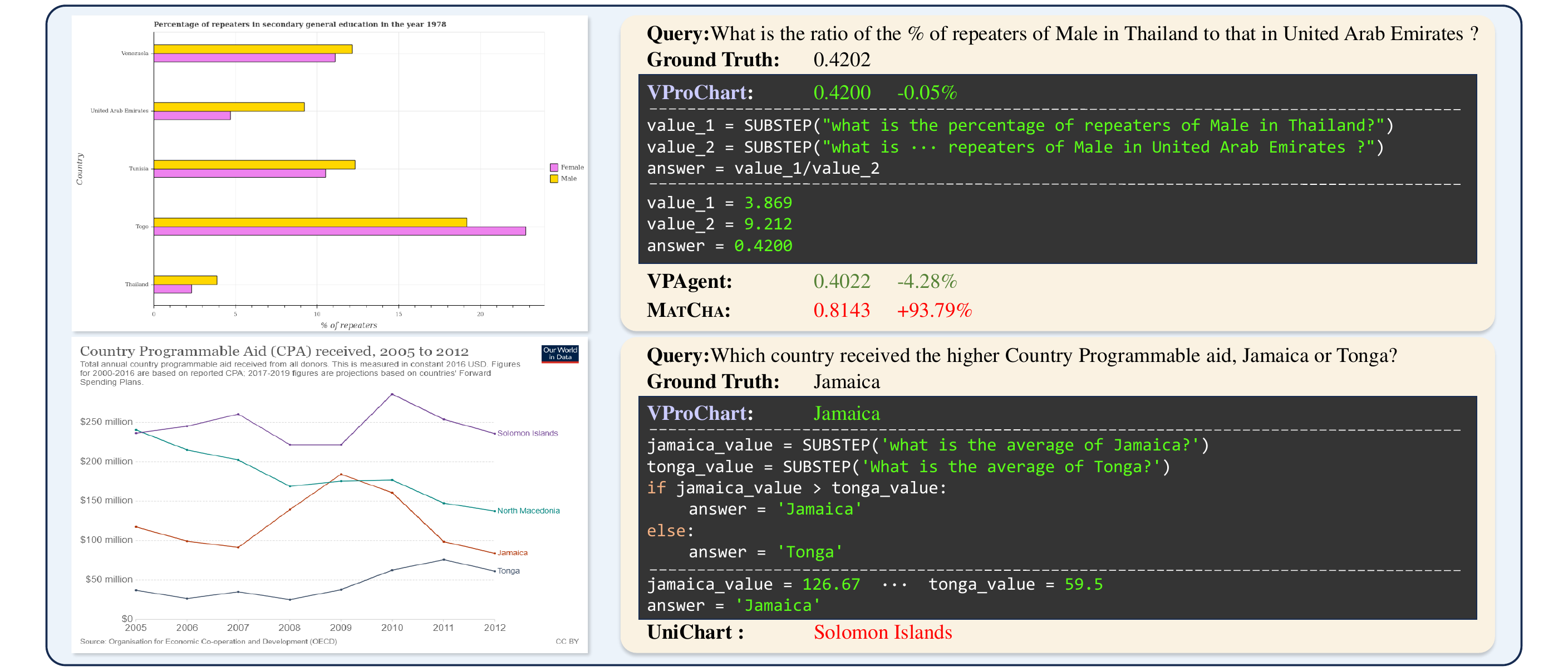}
    \caption{\small Two examples of VProChart. Above and below illustrate two examples from the PlotQA and ChartQA datasets, respectively.}
    \label{case}
\end{figure*}

\begin{table}[ht]
\centering
\begin{tabular}{p{4.1cm}>{\centering\arraybackslash}p{0.8cm}>{\centering\arraybackslash}p{1.2cm}>{\centering\arraybackslash}p{0.5cm}}
\toprule
\multirow{2}{*}{Method}  & \multicolumn{3}{c}{ChartQA} \\ 
\cmidrule(lr){2-4} 
                       & Human & Augment & Avg \\
\midrule
w/o VP alignment       & 44.1  & 87.0    & 65.6 \\
w/o Q-driven chart reasoning & 44.8 & 87.4    & 66.1 \\
w/o intra-cluster reasoning & 46.0 & 87.2    & 66.6 \\
w/o cross-cluster reasoning & 46.6 & 87.4    & 67.0 \\
\rowcolor{gray!20}VPAgent & \textbf{48.1} & \textbf{87.4} & \textbf{67.8} \\
\bottomrule
\end{tabular}
\caption{\small Ablation study on VPAgent.}
\label{Ablation study on VPAgent}
\end{table}

\subsection{Analysis on VPAgent}
\noindent\textbf{VPAgent Ablation.}
To validate the contribution of the VP alignment module and the Q-driven reasoning module within VPAgent, we designed detailed ablation experiments for VPAgent, as shown in Table \ref{Ablation study on VPAgent}. Further experiments, including the zeroshot study and architecture study, can be found in the appendix. We use ``w/o" to indicate the absence of a particular module. It is noteworthy that since VP alignment is a prerequisite module for Q-driven chart reasoning, in the ``w/o VP alignment" experiment, we substitute the alignment matrix $\mathbf{V}$ with Gaussian noise. We can make the following three observations: 
 
1) The performance of models missing any component is significantly lower than VPAgent, demonstrating the effectiveness of our proposed VPAgent for general chart question answering.
    
2) Models with the Q-driven chart reasoning module outperform those without it, highlighting the overall effectiveness of the Q-driven chart reasoning module. However, the model without VP alignment exhibits the lowest performance, indicating the importance of incorporating visual perception into the model.

3) For different versions of the ablation models, the ``human" performance drops much more dramatically than ``Augment". This is reasonable because ``Augment" involves machine-generated images and questions, where answering them does not require the same level of visual complexity as ``human" questions. 
\begin{table}[ht]
\small
\centering
\begin{tabular}{p{2.2cm} >{\centering\arraybackslash}p{2.4cm} >{\centering\arraybackslash}p{2.4cm} }
\toprule
\multirow{2}{*}{Agent Method}  & \multicolumn{2}{c}{ChartQA-Human} \\ 
\cmidrule(lr){2-3} 
                       & w/o Programmatic Solution   & w/ Programmatic Solution      \\ 
\midrule
ChartInstruct-E      & 22.5     & 28.6 $\uparrow$ \textbf{6.1}                 \\
ChartInstruct-P   & 37.4     & 43.2 $\uparrow$ \textbf{5.8}                 \\
ChartGemma        & 28.0     & 34.9 $\uparrow$ \textbf{6.9}                 \\
UniChart        & 43.9      & 52.3 $\uparrow$ \textbf{8.4}                 \\
VPAgent         & 48.1     & 57.4 $\uparrow$ \textbf{9.3}                \\
\bottomrule
\end{tabular}
\caption{\small Ablation study on Programmatic Solution Reasoning. It is noteworthy that all experiments were reproduced locally.}
\label{Programmatic Solution Reasoning Ablation study}
\end{table}

\subsection{Analysis on Programmatic Solution}
\noindent\textbf{Programmatic Solution Ablation.}
In this section, we utilize various chart expert models as agents to investigate the effectiveness of Programmatic Solution Reasoning. We use ``w/o Programmatic Solution” to indicate the performance of the original agent and ``w/ Programmatic Solution” to represent the performance of ChartQA under the guidance of the Programmatic Solution. Considering that the ``Augmented" set almost lacks reasoning issues, our experiments focus on the ChartQA-Human set, which is rich in reasoning problems. The experimental results, as shown in Table \ref{Programmatic Solution Reasoning Ablation study}, indicate that all agents exhibit significant performance improvements under the guidance of the Programmatic Solution. It demonstrate that Programmatic Solution Reasoning is a highly effective method for chart question answering.

\subsection{Case Study}
To further investigate the capabilities of VProChart, we selected sample instances from the PlotQA and ChartQA datasets. Figure \ref{case} illustrates an example from each datasets. In comparison to MatCha or UniChart, our VProChart yields predictions that are closer to the ground truth, even without a Programmatic Solution. This observation underscores the effectiveness of VProChart. Further illustrative examples can be found in the Appendix.

\section{Conclusion}
In this paper, we propose the VProChart, a framework comprising a VPAgent guided by human visual principles and Programmatic Solution Reasoning for precise inference, which presents a novel approach to the ChartQA task. We conducted comprehensive experiments on ChartQA, PlotQA, and DVQA, demonstrating the effectiveness of VProChart, and showing that human visual perception can significantly enhance the model’s capability in understanding schematic diagrams. In the future, we will continue to explore the potential of integrating human-like thinking with programmatic reasoning.

\section*{Acknowledgments}
This work was supported by National Key Research and Development Program of China (2022YFC3303600), the Key Research and Development Project in Shaanxi Province No. 2022GXLH-01-03, National Natural Science Foundation of China (62137002, 62192781, 62293553, 61937001, 62250066, 62306229, and 62106190), ``LENOVO-XJTU'' Intelligent Industry Joint Laboratory Project, Foundation of Key National Defense Science and Technology Laboratory (6142101210201), Natural Science Basic Research Program of Shaanxi (2023-JC-YB-593), the Youth Innovation Team of Shaanxi Universities, the Shaanxi Provincial Social Science Foundation Project (2024P041).

\bibliography{aaai25}
\end{document}